\documentclass{dukecei}
\usepackage[T1]{fontenc}
\usepackage{microtype}
\usepackage{algpseudocode}
\usepackage{array}
\usepackage{multirow}
\usepackage{placeins}
\definecolor{cvprblue}{rgb}{0.21,0.49,0.74}

\title{BATMoS: Budget-Aware Guidance Allocation for Efficient Video Diffusion}

\author{
    Zhixu Du$^{1}$\footnote{Equal contribution. Alphabetical order. Correspondence E-mail: zhixu.du@duke.edu},
    Kuo Yang$^{1}$\footnotemark[1],
    Jingwei Sun$^{2}$,
    Hai "Helen" Li$^{1}$,
    Yiran Chen$^{1}$\\[1em]
    \normalsize $^{1}$Duke University \\
    \normalsize $^{2}$University of Florida\\
}

\begin{document}

\maketitle
\thispagestyle{firstpagestyle}

\begingroup

        \centering
        \includegraphics[width=0.95\textwidth]{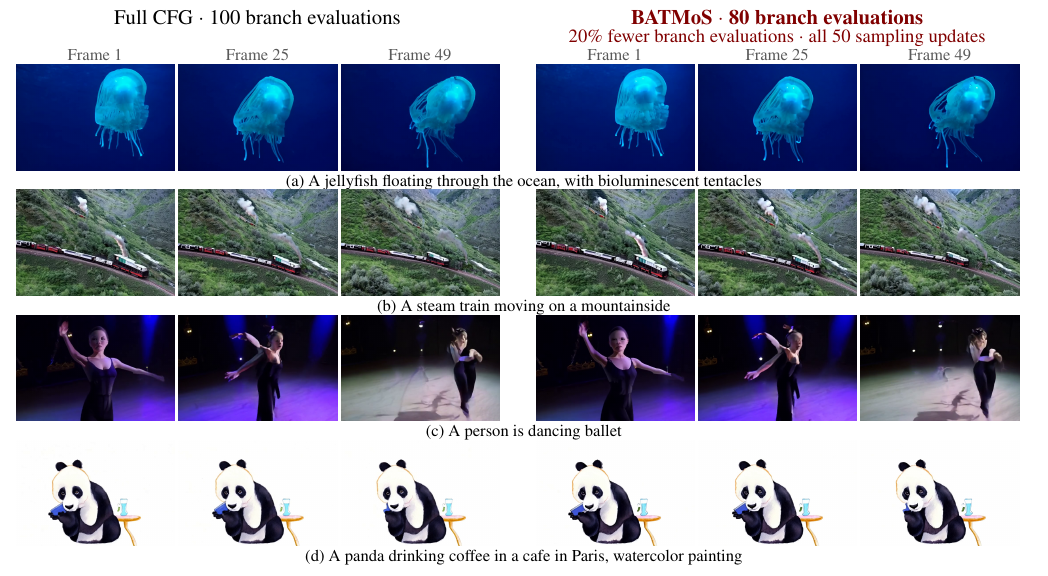}\par
        \captionsetup{hypcap=false}
        \captionof{figure}{%
            \textbf{Video generation with 20\% fewer denoiser branch evaluations.}
            BATMoS retains all 50 sampling updates and selects where to refresh
            guidance through a magnitude-based temporal schedule.
            Four prompts cover marine motion, a moving train, human action,
            and watercolor rendering. Each row compares CogVideoX-2B generations
            under full CFG (left, 100 branches) and BATMoS (right, 80 branches),
            using the same prompt and initial noise.
            Frames 1, 25, and 49 are shown for each video.
        }
        \label{fig:allocation-teaser}
        \par\vspace{12pt}
\endgroup
\begin{abstract}
The effectiveness of a video diffusion model's inference budget depends on where computation is spent along the denoising trajectory. Across eight equal-budget inference configurations, VBench total scores span 4.80 points, highlighting the importance of explicit allocation design. Our key insight is that guidance evaluations are not equally valuable across diffusion time: retaining the sampling updates while selectively reusing guidance can preserve generation quality with fewer denoiser evaluations. We introduce BATMoS (Budget Aware Temporal Mixture of Strategies), which routes between fresh classifier-free guidance and cached guidance using a single pretrained denoiser. A magnitude-based temporal rule assigns guidance reuse to low-magnitude blocks and reserves fresh guidance for the remaining blocks, with a single integer controlling the computation budget. The resulting schedule requires no additional router training or per-budget threshold search. Evaluated on CogVideoX-2B with 4,720 videos per configuration across all sixteen official VBench dimensions, BATMoS achieves the highest total score among the eight allocation policies and comes within 0.10 points of full classifier-free guidance using 20\% fewer branch evaluations. BATMoS also composes with feature reuse, reducing computation from 80 to 62 effective branch evaluations with a 0.11-point decrease in total score. These findings make temporal guidance allocation a practical mechanism for extracting more generation quality from a fixed inference budget.
\end{abstract}

\section{Introduction}
\label{sec:introduction}

Video diffusion models generate a clip through repeated denoising of a spatiotemporal latent, making every additional model evaluation part of the cost of generation~\cite{yang2025cogvideox}. Classifier-free guidance (CFG) compounds this cost by evaluating conditional and unconditional predictions at each guided update~\cite{ho2022classifierfreediffusionguidance}. A 50-update sampler therefore executes 100 denoiser branch evaluations under full CFG. Reducing this budget requires choosing both how many sampling updates to retain and which predictions to recompute along the trajectory.

Prior work provides several mechanisms for reducing fresh computation. Adaptive Guidance omits unconditional evaluations once conditional and unconditional predictions become sufficiently aligned~\cite{castillo2023adaptiveguidancetrainingfreeacceleration}. Restricting guidance to a noise-level interval can improve image generation quality and efficiency~\cite{NEURIPS2024_dd540e1c}. FasterCache exploits temporal feature and CFG-branch redundancy~\cite{lv2025fastercache}, while TeaCache uses timestep-aware signals to schedule network reuse~\cite{Liu_2025_CVPR}. BudCache and BAG make a prescribed budget explicit through offline cache-policy search and learned gating~\cite{lei2026budgetconstrained,zhao2026bagbudgetawaregatingdiffusion}. These approaches expose choices about what to refresh, reuse, or omit~\cite{Du_2025_ICCV}. For guidance allocation, 80 branch evaluations can support 40 fully guided updates, or 50 updates with 30 fully guided updates and 20 cheaper updates.

The sampling grid and the freshness of guidance can be controlled separately. A cheaper update computes a fresh conditional prediction and combines it with a cached conditional–unconditional difference, saving one denoiser branch. FasterCache’s CFG-Cache applies this approximation at a fixed interval~\cite{lv2025fastercache}; we study where along the trajectory it should be applied under a given budget. Guidance magnitude varies along the trajectory, providing a temporal signal for deciding where to use this approximation. We exploit this structure to retain the sampling grid and concentrate guidance refreshes according to the model's guidance-magnitude profile.

We introduce \textbf{BATMoS}, \textbf{Budget Aware Temporal Mixture of Strategies}, which formulates inference as the selection of prediction procedures sharing one pretrained denoiser. Its temporal router partitions the trajectory into blocks and assigns guidance reuse to the blocks with the lowest mean guidance magnitude. The remaining blocks refresh both CFG branches. A single integer specifies how many blocks use cached guidance and directly determines the branch budget. The same fixed block ranking supplies the schedule at each budget, eliminating per-budget threshold search and router training. Figure~\ref{fig:allocation-teaser} shows generated videos using 20\% fewer branch evaluations while retaining all 50 sampling updates.

On CogVideoX-2B, eight inference configurations at this identical budget span \textbf{4.80 VBench total-score points}. BATMoS achieves the highest total score among these allocation policies, at 76.97, compared with 77.07 for full 100-branch CFG. It thus comes within \textbf{0.10 points of full CFG using 20\% fewer branch evaluations}. Each configuration is evaluated on the same 944 prompts with five seeds per prompt, yielding 4,720 videos and covering all sixteen official VBench dimensions~\cite{Huang_2024_CVPR}.

Guidance allocation also complements feature reuse. The two mechanisms approximate different quantities, namely the guidance difference and internal network computation. At 80 effective branch evaluations, BATMoS and TeaCache achieve similar aggregate scores(76.97 and 77.01). BATMoS is higher on imaging quality, while TeaCache is slightly higher on temporal consistency. Composing the magnitude schedule with feature reuse reduces the executed computation from 80 to \textbf{62 effective branch evaluations}, with a 0.11-point decrease in total score.
Our contributions are threefold:
\begin{itemize}
    \item A budgeted temporal mixture of prediction strategies that makes guidance allocation explicit while sharing a single pretrained denoiser.
    \item A magnitude-based allocation rule with direct integer control of the branch budget, achieving the highest score among eight allocation policies and approaching full CFG quality with 20\% fewer branch evaluations.
    \item A composition with feature reuse that reduces computation from 80 to 62 effective branch evaluations with a 0.11-point total-score decrease.
\end{itemize}


\section{Budgeted Temporal Mixture of Strategies}
\label{sec:formulation}

BATMoS allocates computation among prediction procedures that share a pretrained video denoiser. A procedure specifies which predictions to compute and which quantities to retain from earlier updates. The temporal router selects a procedure at each diffusion time under an explicit branch budget.

\subsection{Separate the sampling update from its prediction}

Let $x$ denote a text prompt and $z_0$ the initial noise. A grid of $T$ noise levels $t_0,\ldots,t_{T-1}$ defines the sampling trajectory. At update $i$, a transition $\Phi_i$ consumes a prediction $p_i$ and advances the latent,
\begin{equation}
    z_{i+1}=\Phi_i(z_i,p_i),
    \qquad V=\operatorname{Decode}(z_T).
    \label{eq:transition}
\end{equation}
The prediction uses the parameterization expected by the sampler. In our CogVideoX implementation, this is the model's velocity prediction. The transition and denoiser evaluation are separate operations. A sampler can advance using a fresh prediction, a prediction assembled from fresh and stored quantities, or a stored prediction.

For a frozen denoiser $f_\theta$, write
\begin{equation}
    c_i=f_\theta(z_i,t_i,x),\qquad
    u_i=f_\theta(z_i,t_i,\varnothing),
    \label{eq:branches}
\end{equation}
for conditional and unconditional predictions. With guidance scale $s$, full CFG forms
\begin{equation}
    p_i^{F}=u_i+s(c_i-u_i)=c_i+(s-1)d_i,
    \quad d_i=c_i-u_i.
    \label{eq:cfg}
\end{equation}
The second expression exposes two quantities with different refresh requirements. The conditional prediction responds to the current latent. The guidance difference supplies the additional conditional correction. Evaluating both branches refreshes both quantities at a cost of two branch evaluations.

\subsection{Prediction strategies with shared parameters}

Let $r<i$ denote the most recent update that refreshed the guidance difference. A guidance-reuse update computes $c_i$ and forms
\begin{equation}
    p_i^{G}=c_i+(s-1)d_r.
    \label{eq:guidance-reuse}
\end{equation}
This update spends one fresh branch while retaining a model-driven prediction of the current latent. It requires one stored tensor with the same shape as a denoiser prediction. A full CFG update replaces this tensor with $d_i$. BATMoS initializes the trajectory with full CFG, so the guidance cache is populated before reuse.

Table~\ref{tab:strategies} defines the strategy vocabulary. BATMoS's magnitude schedule selects between $F$ and $G$. Conditional-only updates $C$ and prediction reuse $P$ describe the other allocation policies in the evaluation. Each procedure invokes the same denoiser parameters. Their specialization concerns the quantities refreshed at a particular diffusion time.

\begin{table}[t]
    \centering
    \small
    \caption{\textbf{Prediction procedures sharing one denoiser.} Costs count fresh full-stack branch evaluations. $d_r$ is the stored guidance difference and $p_{i-1}$ the previous prediction. Every procedure supplies a prediction to the sampler transition.}
    \label{tab:strategies}
    \begin{tabular}{@{}llc@{}}
        \toprule
        Procedure & Prediction & Branch cost \\
        \midrule
        Full CFG ($F$) & $c_i+(s-1)d_i$ & 2 \\
        Guidance reuse ($G$) & $c_i+(s-1)d_r$ & 1 \\
        Conditional ($C$) & $c_i$ & 1 \\
        Prediction reuse ($P$) & $p_{i-1}$ & 0 \\
        \bottomrule
    \end{tabular}
\end{table}

\subsection{Allocate a branch budget}

A strategy sequence $\pi=(a_0,\ldots,a_{T-1})$ generates a video $V_\pi$ through Eq.~\eqref{eq:transition}. Let $n(a_i)$ be its branch cost and $Q$ a completed-video utility. The budgeted selection problem is
\begin{equation}
    \max_{\pi\in\Pi}\ \mathbb{E}_{x,z_0}[Q(V_\pi)]
    \quad\text{subject to}\quad
    \sum_{i=0}^{T-1}n(a_i)\leq B.
    \label{eq:budget-objective}
\end{equation}
Here $\Pi$ specifies the strategy vocabulary, initialization, and temporal structure. A fixed temporal router selects $a_i$ from the update index and the requested budget. The final-video objective accounts for the propagation of prediction changes through the remaining trajectory.

For an $F/G$ schedule, every update evaluates at least one branch. If $M$ updates use full CFG, the total is
\begin{equation}
    B=2M+(T-M)=T+M.
    \label{eq:budget-count}
\end{equation}
Thus, an 80-branch budget on a 50-update grid provides 30 full CFG updates and 20 guidance-reuse updates. A 40-update full-CFG sampler spends the same 80 branches on a different grid. The allocation problem determines how to distribute the available fresh predictions over the trajectory.

\paragraph{Computation units.}
Two branches count as two evaluations even when a single batched call executes them. Internal feature reuse can further reduce transformer computation within a call. For a denoiser with $L$ transformer blocks, we report
\begin{equation}
    N_{\mathrm{eff}}=\frac{1}{L}
    \sum_i\sum_{\ell=1}^{L} b_{i,\ell},
    \label{eq:effective-nfe}
\end{equation}
where $b_{i,\ell}$ is the number of branch elements actually processed by block $\ell$ at update $i$. A complete single-branch evaluation contributes one effective NFE. For $F/G$ schedules that execute every block, $N_{\mathrm{eff}}=B$. This unit also accounts for the block computation removed by feature reuse. Input and output projections remain part of the implementation and its runtime.

\section{Magnitude-Based Temporal Allocation}
\label{sec:method}

BATMoS constructs a reusable ordering of diffusion-time blocks from the denoiser's guidance-magnitude profile. The requested budget determines how many blocks receive guidance reuse. The resulting schedule keeps the sampling grid fixed and gives each budget an explicit execution sequence.

\subsection{Read the guidance-magnitude profile}

Under ordinary full-CFG sampling, define the normalized guidance magnitude
\begin{equation}
    m_i=\frac{\lVert c_i-u_i\rVert_2}{\lVert c_i\rVert_2}.
    \label{eq:magnitude}
\end{equation}
The norm spans the prediction tensor. This quantity measures the size of the guidance difference relative to the current conditional prediction. Its temporal variation provides an ordering for allocating guidance refreshes.

We collect the profile from six prompts and two seeds, giving twelve full-CFG trajectories. Partition the $T=50$ updates into $J=10$ contiguous blocks of width $h=5$. For block $I_j$, its score is
\begin{equation}
    \bar m_j=\frac{1}{Nh}\sum_{n=1}^{N}
    \sum_{i\in I_j}m_i^{(n)},\qquad N=12.
    \label{eq:block-score}
\end{equation}
We construct the ranking directly from these predictions, without completed-video scoring or router training, and reuse it across the evaluated prompts, seeds, and requested budgets. The reference prompts and per-block values are given in the Supplementary Material.

\begin{figure*}[t]
    \centering
    \includegraphics[width=\textwidth]{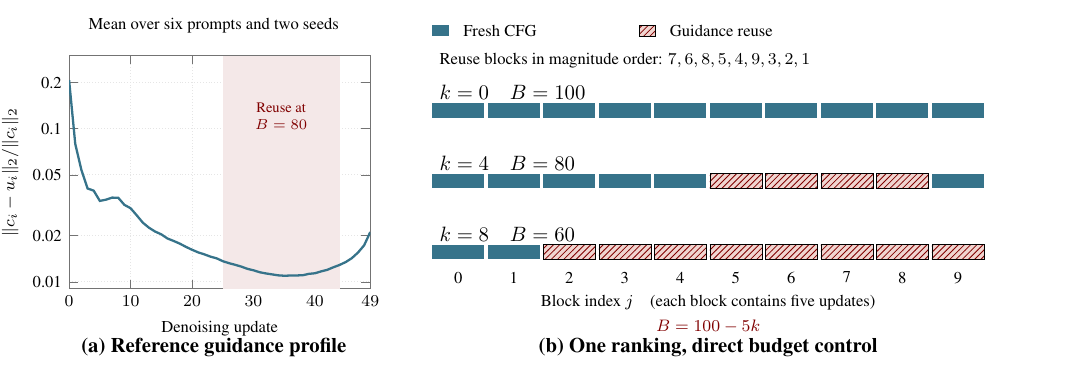}
    \caption{\textbf{A guidance-magnitude profile gives a direct temporal budget control.}
    (a) Mean normalized guidance magnitude from twelve ordinary full-CFG trajectories on CogVideoX-2B. The vertical axis is logarithmic. The shaded region contains the blocks selected for guidance reuse at $B=80$.
    (b) Replacing one five-update block of fresh CFG with guidance reuse saves five branch evaluations. A prefix of the fixed magnitude ranking determines the schedule, with $B=100-5k$.}
    \label{fig:magnitude-rule}
\end{figure*}

The measured ordering, from smallest to largest block magnitude, is
\begin{equation}
    (7,6,8,5,4,9,3,2,1,0).
    \label{eq:ranking}
\end{equation}
The low-magnitude region lies late in the trajectory, followed by an increase in the final block. This profile allocates fresh guidance to the beginning and end of the deployed 80-branch schedule, with a contiguous guidance-reuse region between them.

\subsection{Map an integer directly to computation}

Let $\mathcal G_k$ contain the $k$ lowest-ranked reusable blocks. The initial block uses full CFG to initialize the guidance cache. For $k\leq J-1$, the temporal router is
\begin{equation}
    a_i=
    \begin{cases}
        G,& \lfloor i/h\rfloor\in\mathcal G_k,\\
        F,& \text{otherwise}.
    \end{cases}
    \label{eq:temporal-rule}
\end{equation}
Each reused block replaces $h$ two-branch updates with $h$ one-branch updates. Its exact branch count is therefore
\begin{equation}
    B(k)=2T-hk=100-5k.
    \label{eq:integer-budget}
\end{equation}
The integer $k$ directly selects the computation budget. The block ordering remains fixed as $k$ changes, so selecting another budget requires taking a longer or shorter prefix of that ordering.

At $k=4$, blocks $5$ through $8$ use guidance reuse. Writing exponents for repetitions, the deployed sequence is
\begin{equation}
    F^{25}G^{20}F^5,
    \qquad B(4)=80.
    \label{eq:deployed}
\end{equation}
This schedule refreshes guidance at 30 updates and retains a fresh conditional prediction at all 50. At $k=8$, the same rule produces $F^{10}G^{40}$ and a 60-branch budget. Figure~\ref{fig:magnitude-rule} connects the measured profile, block selection, and these exact budgets.

\paragraph{Cache evolution along the schedule.}
For $F^{25}G^{20}F^5$, update 24 writes $d_{24}$ to the guidance cache. Updates 25--44 each evaluate their own conditional prediction and combine it with this stored difference. Update 45 recomputes both branches and replaces the cache with $d_{45}$. The final five updates therefore resume fresh guidance. In $F^{10}G^{40}$, update 9 supplies the difference used by all subsequent updates. The budget changes the duration of guidance reuse while every update continues to respond to the evolving latent through $c_i$.

\paragraph{Offline ordering and online execution.}
The ten-block ranking is computed once from the reference trajectories. A budget request selects its prefix and expands the block assignments into a 50-entry strategy sequence. Blocks give a five-branch budget resolution; reserving the initial block for CFG admits $0\leq k\leq9$. During sampling, routing is an index lookup. Full-CFG updates batch the unconditional and conditional inputs into one two-branch call, while guidance-reuse updates use a one-branch call. The guidance cache holds one prediction-shaped tensor per active trajectory, with no per-layer guidance history.

\begin{algorithm}[t]
    \caption{BATMoS temporal allocation and sampling}
    \label{alg:batmos}
    \begin{algorithmic}[1]
        \Require Frozen denoiser $f_\theta$, prompt $x$, initial noise $z_0$, fixed block ranking $\rho$, budget $B=2T-hk$
        \State Select the first $k$ reusable blocks in $\rho$ as $\mathcal G_k$
        \State $d_{\mathrm{cache}}\gets\emptyset$\quad(first block is assigned to $F$)
        \For{$i=0,\ldots,T-1$}
            \State Compute $c_i=f_\theta(z_i,t_i,x)$
            \If{$\lfloor i/h\rfloor\notin\mathcal G_k$}
                \State Compute $u_i=f_\theta(z_i,t_i,\varnothing)$
                \State $d_{\mathrm{cache}}\gets c_i-u_i$
            \EndIf
            \State $p_i\gets c_i+(s-1)d_{\mathrm{cache}}$
            \State $z_{i+1}\gets\Phi_i(z_i,p_i)$
        \EndFor
        \State \Return $\operatorname{Decode}(z_T)$
    \end{algorithmic}
\end{algorithm}

\paragraph{What the approximation changes.}
At a guidance-reuse update, the conditional prediction is evaluated at the current latent. The local prediction error relative to full CFG at that same latent is
\begin{equation}
    p_i^G-p_i^F=(s-1)(d_r-d_i).
    \label{eq:reuse-error}
\end{equation}
Guidance reuse therefore preserves the conditional response to the current latent while approximating the evolution of the guidance difference. Magnitude supplies the temporal allocation signal, and $d_r-d_i$ determines the local prediction error.

\subsection{Compose guidance allocation with feature reuse}

Guidance reuse and feature reuse operate at different levels of the prediction procedure. Guidance reuse carries a conditional--unconditional difference between updates. Feature reuse carries internal network residuals between evaluations. BATMoS combines them by applying a TeaCache-style residual gate inside contiguous runs of CFG updates~\cite{Liu_2025_CVPR}.

\begin{figure*}[t]
    \centering
    \includegraphics[width=\textwidth]{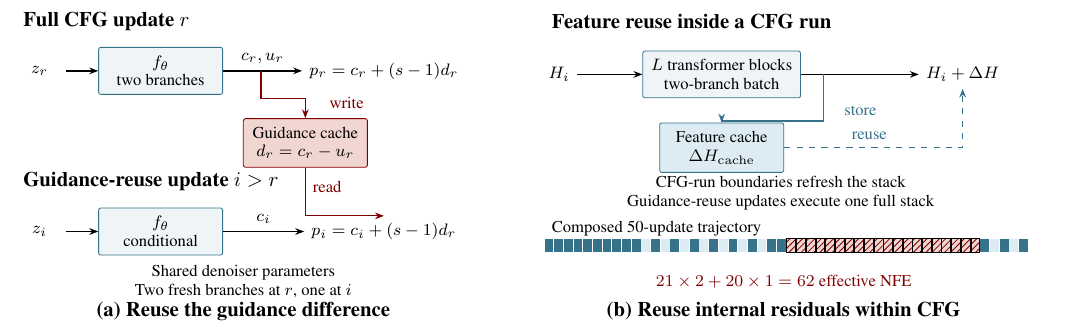}
    \caption{\textbf{Guidance allocation and feature reuse retain different quantities.}
    (a) Full CFG refreshes the conditional--unconditional difference. A later guidance-reuse update combines that cached difference with a fresh conditional prediction.
    (b) Internal residual reuse operates inside two-branch CFG runs. Run boundaries refresh the stack, and one-branch guidance-reuse updates execute all blocks. The strip shows full-stack CFG (dark blue), guidance reuse (hatched red), and residual reuse (pale blue). Their update counts are 21, 20, and nine, yielding 62 effective NFE.}
    \label{fig:two-caches}
\end{figure*}

The cache contract follows the branch shape of each update. A full-CFG call processes a two-branch batch and can populate the feature residual cache. A guidance-reuse call processes a single conditional branch and executes the full transformer stack. Each CFG run starts and ends with a full-stack refresh. This rule gives the feature cache a consistent two-branch shape and keeps the stored guidance difference available for the subsequent one-branch updates. Figure~\ref{fig:two-caches} illustrates these two retained quantities.

At a full-stack CFG update $q$, the feature cache stores the transformer's output-minus-input residuals for both video and text tokens. A residual-reuse update adds these tensors to its current token embeddings, then executes the final normalization and output projection. Its two output branches form a new guidance difference for the outer cache. A single-branch guidance-reuse update leaves the two-branch feature cache untouched. Thus the inner cache reduces the work required to produce CFG branches, while the outer cache reduces how often the unconditional branch is requested.

\paragraph{Residual gate within a CFG run.}
Let $e_i$ be the timestep embedding and $e_{\mathrm{prev}}$ its value at the preceding CFG call in the run. The gate computes a relative embedding change and accumulates its polynomially rescaled value,
\begin{equation}
    q_i=\frac{\lVert e_i-e_{\mathrm{prev}}\rVert_1}{\lVert e_{\mathrm{prev}}\rVert_1},
    \qquad A\gets A+P(q_i).
    \label{eq:residual-gate}
\end{equation}
The polynomial $P$ follows the CogVideoX-2B TeaCache implementation. If $A<\tau$, the call reuses the residuals; otherwise it refreshes the full stack and resets $A$ to zero. Run boundaries force a refresh and reset, so the two CFG regions have independent accumulation histories. This gate uses timestep embeddings, giving the fixed model and grid a deterministic reuse pattern. The polynomial coefficients and reuse indices are provided in the Supplementary Material.

For the schedule in Eq.~\eqref{eq:deployed}, the CFG runs occupy updates $0$--$24$ and $45$--$49$. With residual threshold $\tau=0.1$, nine of these 30 CFG updates reuse internal residuals. The other 21 evaluate both branches through the full stack, and all 20 guidance-reuse updates evaluate one branch through the full stack. Consequently,
\begin{equation}
    N_{\mathrm{eff}}=21\times2+20\times1=62.
    \label{eq:composed-cost}
\end{equation}
All 50 sampler updates remain in the trajectory. The base schedule determines where guidance is refreshed, and the residual gate reduces the internal computation of selected CFG updates. The measured quality and cost of this composition are reported with the individual mechanisms in Sec.~\ref{sec:composition}.

\section{Evaluation}
\label{sec:evaluation}

We evaluate temporal allocation at a fixed branch budget, quality relative to full CFG, and composition with feature reuse. The principal operating point is the 80-branch schedule in Eq.~\eqref{eq:deployed}.

\subsection{Experimental setup}

\paragraph{Model and population.}
We use CogVideoX-2B~\cite{yang2025cogvideox} at $480\times720$ resolution with 49 frames and guidance scale $s=6$. The sampler is CogVideoX DDIM with $\eta=0$. The common grid contains 50 updates, with a 40-update full-CFG configuration providing the fewer-update baseline. Each configuration generates videos for the same 944 unique VBench prompts and five seeds, yielding 4,720 videos per configuration. This population covers all sixteen official VBench dimensions~\cite{Huang_2024_CVPR}. The evaluation contains fourteen configurations across the allocation, feature-reuse, and composition comparisons.

\paragraph{Scores and computation.}
We report official VBench total, quality, and semantic scores on a 0--100 scale. Dimension scores are the raw dimension aggregates multiplied by 100. Score differences are in points. The computation unit is effective NFE from Eq.~\eqref{eq:effective-nfe}, measured from executed transformer blocks. For full-stack guidance-allocation policies this equals their branch budget. Generation runs on NVIDIA A100-SXM4-80GB GPUs with BF16 model weights and latents. Joint-branch configurations use FP32 prediction, guidance, and sampler arithmetic. The compact-interval, prior-based, and stated-based selectors use separate branch calls and the BF16 prediction arithmetic of their calibration. Table~\ref{tab:allocation-main} identifies these configurations, and the Supplementary Material specifies their execution and initialization.

\paragraph{Policy construction.}
BATMoS uses the fixed profile and ranking in Sec.~\ref{sec:method}. Reference and calibration prompts are drawn from the benchmark pool, and policies are fixed before full-population evaluation. The Supplementary Material provides the prompt selections, policy sequences, and feature-reuse gate parameters.

\paragraph{Compared policies.}
  Using the strategy notation in Table~1, $F$, $G$, $C$, and $P$ denote full CFG, guidance reuse, conditional prediction, and previous-prediction reuse, respectively; superscripts indicate repeated updates. Front-loaded guidance uses $F^{30}C^{20}$, the shifted guidance interval uses $C^5F^{30}C^{15}$, and semantic-window allocation uses $F^{10}G^5(F^5C^5)^3F^5$.
  The compact-interval selector uses $FC^9F^{29}C^{11}$. The prior-based and state-based selectors modify this compact-interval schedule by selecting equal-cost four-update patterns from $\{C^4,G^4,(FP)^2\}$. The prior-based selector uses window-wise calibration mean utilities; the state-based selector augments these means with a ridge-regression correction from
  trajectory features. The 40-update CFG baseline evaluates both branches on a shorter sampling grid, while full CFG uses all 50 updates. TeaCache reuses internal transformer residuals, with its threshold selected to match 80 effective NFE.

\subsection{Allocation changes quality at a fixed budget}
\label{sec:allocation-results}

\begin{table*}[t]
\centering\small
\caption{\textbf{Allocation changes video quality at a fixed computation budget.} Eight guidance-allocation policies execute 80 full-stack branch evaluations. BATMoS has the highest total score among these policies. The reference rows report TeaCache at matched effective NFE and full 100-branch CFG. $^{\dagger}$ marks separate-branch execution with BF16 prediction arithmetic. The remaining configurations use joint-branch execution with FP32 prediction arithmetic. All scores use the same 4,720-video population and official VBench aggregation.}
\label{tab:allocation-main}
\begin{tabular}{@{}lrrrrr@{}}
\toprule
Method & Sampling updates & Effective NFE & Total & Quality & Semantic \\
\midrule
\textbf{BATMoS} & 50 & 80 & 76.97 & 79.40 & 67.25 \\
Front-loaded guidance & 50 & 80 & 76.86 & 79.32 & 66.99 \\
Prior-based selector$^{\dagger}$ & 50 & 80 & 76.67 & 79.09 & 66.96 \\
Shifted guidance interval & 50 & 80 & 76.57 & 79.68 & 64.15 \\
Semantic-window allocation & 50 & 80 & 76.40 & 78.97 & 66.12 \\
40-update CFG & 40 & 80 & 76.32 & 78.88 & 66.11 \\
State-based selector$^{\dagger}$ & 50 & 80 & 76.16 & 78.66 & 66.16 \\
Compact-interval selector$^{\dagger}$ & 50 & 80 & 72.17 & 76.08 & 56.52 \\
\midrule
TeaCache, matched budget & 50 & 80 & 77.01 & 79.40 & 67.47 \\
Full CFG & 50 & 100 & 77.07 & 79.45 & 67.57 \\
\bottomrule
\end{tabular}
\end{table*}

The eight 80-branch allocation policies in Table~\ref{tab:allocation-main} span \textbf{4.80 total-score points}, from 72.17 to 76.97. BATMoS achieves the highest total score among these policies.

Its advantage over the other joint-execution allocations ranges from 0.11 points for front-loaded guidance to 0.65 points for 40-update CFG. Both BATMoS and 40-update CFG execute 80 full-stack branches, while BATMoS distributes them over 50 updates and scores 76.97 against 76.32. The magnitude schedule also scores above the shifted guidance interval and semantic-window allocation by 0.40 and 0.57 points, respectively. Relative to the shifted interval, its semantic score increases from 64.15 to 67.25.

\paragraph{20\% fewer branch evaluations.}
Against the 100-branch reference, BATMoS reduces branch evaluations by \textbf{20\%} and changes total score from 77.07 to 76.97. Quality scores are 79.45 and 79.40, and semantic scores are 67.57 and 67.25. These results locate the 80-branch schedule close to full CFG across both aggregate components. Figure~\ref{fig:allocation-teaser} shows four paired generations at matched prompts and initial noise, spanning natural scenes, articulated motion, and stylized content.

\subsection{Feature reuse extends the 80-branch schedule}
\label{sec:composition}

\begin{table*}[t]
\centering\small
\caption{\textbf{Feature reuse extends the 80-branch BATMoS schedule to 62 effective NFE.} The execution columns count sampler updates with full-stack CFG, full-stack guidance reuse, or CFG residual reuse. Every row retains 50 sampling updates. The complete operating range includes the 60-branch magnitude schedule and TeaCache's lower-cost configurations.}
\label{tab:range-main}
\begin{tabular}{@{}lrrrrrrr@{}}
\toprule
Method & Eff. NFE & \shortstack{Full-stack\\CFG} & \shortstack{Guidance\\reuse} & \shortstack{Residual\\reuse} & Total & Quality & Semantic \\
\midrule
Full CFG & 100 & 50 & 0 & 0 & 77.07 & 79.45 & 67.57 \\
\textbf{BATMoS} & 80 & 30 & 20 & 0 & 76.97 & 79.40 & 67.25 \\
\textbf{BATMoS + feature reuse} & 62 & 21 & 20 & 9 & 76.86 & 79.30 & 67.09 \\
\textbf{BATMoS} & 60 & 10 & 40 & 0 & 72.51 & 75.94 & 58.75 \\
\midrule
TeaCache, matched budget & 80 & 40 & 0 & 10 & 77.01 & 79.40 & 67.47 \\
TeaCache, matched budget & 60 & 30 & 0 & 20 & 76.88 & 79.26 & 67.37 \\
TeaCache ($\tau=0.1$) & 54 & 27 & 0 & 23 & 76.82 & 79.26 & 67.07 \\
\bottomrule
\end{tabular}
\end{table*}

The composed schedule executes \textbf{62 effective NFE} and scores 76.86, compared with 80 effective NFE and 76.97 for BATMoS alone. It removes \textbf{22.5\%} of the 80-branch schedule's block computation with a 0.11-point change in total score. Its quality score is 79.30 and its semantic score is 67.09. The execution count follows the 21 full-stack CFG updates and 20 full-stack guidance-reuse updates in Eq.~\eqref{eq:composed-cost}. Nine additional CFG updates reuse internal residuals while the sampler advances.

\paragraph{Preserving guidance opportunities at lower cost.}
Reducing the magnitude-rule budget to 60 selects $F^{10}G^{40}$ and scores 72.51. The 62-NFE composition scores 4.35 points higher while retaining the 80-branch schedule's 30 guidance-enabled updates. Twenty-one of these updates refresh the full stack and nine use cached residuals. Feature reuse thus reduces computation within selected CFG evaluations while preserving their positions in the guidance schedule. Table~\ref{tab:range-main} reports the complete operating range alongside the feature-reuse references.

\subsection{The approximations affect different video attributes}
\label{sec:dimension-results}

\begin{table}[t]
\centering\small
\caption{\textbf{Guidance allocation and feature reuse affect different attributes.} Frame-quality and temporal-consistency dimensions at 80 effective NFE. $\Delta$ is BATMoS minus TeaCache in raw dimension-score points. The Supplementary Material reports all sixteen dimensions.}
\label{tab:dimensions-main}
\begin{tabular}{@{}lrrr@{}}
\toprule
Dimension & BATMoS & TeaCache & $\Delta$ \\
\midrule
Imaging quality & 57.82 & 57.24 & $+0.58$ \\
Temporal flickering & 96.80 & 96.84 & $-0.04$ \\
Motion smoothness & 96.92 & 97.02 & $-0.10$ \\
Subject consistency & 93.23 & 93.36 & $-0.13$ \\
\bottomrule
\end{tabular}
\end{table}

At 80 effective NFE, BATMoS and TeaCache both round to 79.40 on quality. BATMoS improves imaging quality by \textbf{0.58 points}, while TeaCache is 0.04–0.13 points higher on the three temporal-consistency dimensions. Table~\ref{tab:dimensions-main} reports the frame-quality and temporal-consistency dimensions, with the complete sixteen-dimension comparison in the Supplementary Material.

Guidance reuse refreshes the conditional prediction of the current latent at every update. Residual reuse approximates internal network computation while preserving the CFG procedure. The dimension-level results make these different approximations visible even when the aggregate quality scores are close. Their practical compatibility is established by the 62-NFE composition.

\subsection{Computation accounting and modeled time}
\label{sec:timing}

The executed block counts are 3,000 branch-block elements for full CFG, 2,400 for BATMoS, and 1,860 for the composition. Dividing by the 30 transformer blocks yields 100, 80, and 62 effective NFE. An affine model predicts generation times of 86.0, 71.4, and 58.2 seconds, respectively, including prompt preparation, denoising, and VAE decoding. The Supplementary Material gives the model, its fit inputs, and the observed shared-server timings.

\section{Related Work}
\label{sec:related}

\paragraph{Video generative models.}
Denoising diffusion and score-based models generate samples through successive predictions~\cite{NEURIPS2020_4c5bcfec,song2021scorebased}, with preconditioning and sampling characterized by EDM~\cite{NEURIPS2022_a98846e9}. Latent diffusion reduces spatial computation~\cite{Rombach_2022_CVPR}, while DiT scales transformer backbones~\cite{Peebles_2023_ICCV}; flow matching and rectified flow learn continuous transport fields~\cite{lipman2023flowmatchinggenerativemodeling,liu2022flowstraightfastlearning}. Video Diffusion Models and Imagen Video model spatiotemporal denoising~\cite{NEURIPS2022_39235c56,ho2022imagenvideohighdefinition}. Image-to-video transfer supports temporal generation~\cite{singer2022makeavideotexttovideogenerationtextvideo,Blattmann_2023_CVPR,guo2024animatediff}. Latte, CogVideoX, HunyuanVideo, and Wan develop transformer-based video generators~\cite{ma2025latte,yang2025cogvideox,kong2025hunyuanvideosystematicframeworklarge,wan2025wanopenadvancedlargescale}.

\paragraph{Fewer-step generation.}
DDIM and DPM-Solver reduce sampling evaluations through deterministic trajectories and numerical integration~\cite{song2022denoisingdiffusionimplicitmodels,NEURIPS2022_260a14ac}. Progressive distillation learns to reproduce multiple teacher updates in fewer student steps~\cite{salimans2022progressivedistillationfastsampling}. Consistency models and latent consistency models learn direct mappings along generation trajectories~\cite{pmlr-v202-song23a,luo2023latentconsistencymodelssynthesizing}. Adversarial and distribution-matching distillation train generators for few-step synthesis~\cite{sauer2023adversarialdiffusiondistillation,Yin_2024_CVPR}.

\paragraph{Guidance computation.}
Classifier guidance and CFG control the generated distribution through additional conditional signals~\cite{NEURIPS2021_49ad23d1,ho2022classifierfreediffusionguidance}. Adaptive Guidance omits unconditional evaluations as branch directions align~\cite{castillo2023adaptiveguidancetrainingfreeacceleration}, and limited-interval guidance varies where guidance is applied~\cite{NEURIPS2024_dd540e1c}. FasterCache exploits temporal feature redundancy together with CFG-branch redundancy~\cite{lv2025fastercache}.

\paragraph{Caching granularity.}
Caching reduces repeated work through reusable internal representations~\cite{Ma_2024_CVPR,Wimbauer_2024_CVPR} and manages memory through predicted access to sparse expert parameters~\cite{MLSYS2024_698cfaf7}. FORA caches attention and MLP outputs~\cite{selvaraju2024forafastforwardcachingdiffusion}, while $\Delta$-DiT reuses block residuals across denoising stages~\cite{chen2024deltadittrainingfreeaccelerationmethod}. PAB broadcasts attention outputs over module-specific intervals~\cite{ICLR2025_092c2d45}; TGATE exploits temporal changes in attention roles~\cite{liu2025fasterdiffusiontemporalattention}; ToCa selects tokens according to caching sensitivity~\cite{zou2025accelerating}. BATMoS reuses a cached conditional–unconditional difference, as in FasterCache’s CFG-Cache~\cite{lv2025fastercache}, and addresses where to place this reuse under a budget.

\paragraph{Cache prediction and correction.}
TeaCache uses timestep-aware signals to schedule residual reuse~\cite{Liu_2025_CVPR}, and AdaCache adapts reuse to video denoising behavior~\cite{Kahatapitiya_2025_ICCV}. TaylorSeer and LESA predict future features through temporal expansion and learned stage-specific predictors~\cite{Liu_2025_ICCV,Cai_2026_CVPR}. Increment-calibrated caching and gradient-optimized caching correct retained features~\cite{Chen_2025_CVPR,Qiu_2025_ICCV}. ARCache uses generation history and residual correction for autoregressive video diffusion~\cite{Nan_2026_CVPR}, while TC-Pad\'e develops trajectory-consistent rational approximation~\cite{He_2026_CVPR}.

\paragraph{Budgeted allocation.}
Compute-optimal deployment jointly considers denoising steps and per-step inference cost~\cite{Du_2025_ICCV}. In expert-serving systems, prediction granularity and load balancing jointly determine the cost of routed computation~\cite{ma2026moegps}. Learning-to-Cache optimizes layer-wise caching decisions~\cite{ma2024learningtocache}; MagCache schedules feature reuse from residual-magnitude behavior~\cite{ma2025magcache}. BudCache searches a cache mask under a prescribed budget~\cite{lei2026budgetconstrained}, BAG learns budget-conditioned gating~\cite{zhao2026bagbudgetawaregatingdiffusion}, and GCache optimizes reuse through its effect on final quality~\cite{ye2026localmismatchglobalimpact}. MixCache coordinates multiple caching granularities~\cite{wei2026adaptivehybridcachingefficient}. BATMoS constructs a guidance-difference schedule from a fixed magnitude ordering and an exact block-cost relation, without completed-video scoring during schedule construction.

\paragraph{Conditional computation.}
Expert mixtures allocate model capacity through sparse language-model routing~\cite{shazeer2017outrageouslylargeneuralnetworks,JMLR:v23:21-0998}, visual routing~\cite{NEURIPS2021_48237d9f}, differentiable expert mixtures~\cite{puigcerver2024from}, and state-space parameter mixtures with a shared recurrent trajectory~\cite{du2026swimbaswitchmambamodel}. 
BATMoS applies temporal selection to prediction strategies that share one frozen denoiser, where the requested budget controls its execution schedule.


\section{Conclusion}
\label{sec:conclusion}

BATMoS makes temporal guidance allocation explicit under a denoiser branch budget. A fixed magnitude-based ranking selects where to reuse guidance, retaining fresh conditional predictions throughout the sampling grid. Across eight inference configurations at an 80-branch budget, total scores span 4.80 points, with BATMoS achieving the highest score among the allocation policies. Its deployed schedule comes within 0.10 points of full CFG using 20\% fewer branch evaluations. Composing this schedule with feature reuse reduces computation from 80 to 62 effective NFE with a 0.11-point score change. Temporal guidance allocation provides a direct budget control and a complementary way to reduce video diffusion computation.
\label{sec:conclusion-end}

\clearpage
\section*{LLM Usage Statement}
OpenAI Codex assisted with manuscript drafting and editing, literature-source checking, LaTeX figure and table preparation, and consistency review. The quantitative results and illustrated video frames are taken from the recorded experimental data.

{
    \small
    \bibliographystyle{abbrvnat}
    \bibliography{references}

@misc{ho2022classifierfreediffusionguidance,
      title={Classifier-Free Diffusion Guidance}, 
      author={Jonathan Ho and Tim Salimans},
      year={2022},
      eprint={2207.12598},
      archivePrefix={arXiv},
      primaryClass={cs.LG},
      url={https://arxiv.org/abs/2207.12598}, 
}

@inproceedings{
yang2025cogvideox,
title={CogVideoX: Text-to-Video Diffusion Models with An Expert Transformer},
author={Zhuoyi Yang and Jiayan Teng and Wendi Zheng and Ming Ding and Shiyu Huang and Jiazheng Xu and Yuanming Yang and Wenyi Hong and Xiaohan Zhang and Guanyu Feng and Da Yin and Yuxuan.Zhang and Weihan Wang and Yean Cheng and Bin Xu and Xiaotao Gu and Yuxiao Dong and Jie Tang},
booktitle={The Thirteenth International Conference on Learning Representations},
year={2025},
url={https://openreview.net/forum?id=LQzN6TRFg9}
}

@misc{castillo2023adaptiveguidancetrainingfreeacceleration,
      title={Adaptive Guidance: Training-free Acceleration of Conditional Diffusion Models}, 
      author={Angela Castillo and Jonas Kohler and Juan C. Pérez and Juan Pablo Pérez and Albert Pumarola and Bernard Ghanem and Pablo Arbeláez and Ali Thabet},
      year={2023},
      eprint={2312.12487},
      archivePrefix={arXiv},
      primaryClass={cs.LG},
      url={https://arxiv.org/abs/2312.12487}, 
}

@inproceedings{NEURIPS2024_dd540e1c,
 author = {Kynk\"{a}\"{a}nniemi, Tuomas and Aittala, Miika and Karras, Tero and Laine, Samuli and Aila, Timo and Lehtinen, Jaakko},
 booktitle = {Advances in Neural Information Processing Systems},
 doi = {10.52202/079017-3892},
 editor = {A. Globerson and L. Mackey and D. Belgrave and A. Fan and U. Paquet and J. Tomczak and C. Zhang},
 pages = {122458--122483},
 publisher = {Curran Associates, Inc.},
 title = {Applying Guidance in a Limited Interval Improves Sample and Distribution Quality in Diffusion Models},
 url = {https://proceedings.neurips.cc/paper_files/paper/2024/file/dd540e1c8d26687d56d296e64d35949f-Paper-Conference.pdf},
 volume = {37},
 year = {2024}
}

@inproceedings{
lv2025fastercache,
title={FasterCache: Training-Free Video Diffusion Model Acceleration with High Quality},
author={Zhengyao Lv and Chenyang Si and Junhao Song and Zhenyu Yang and Yu Qiao and Ziwei Liu and Kwan-Yee K. Wong},
booktitle={The Thirteenth International Conference on Learning Representations},
year={2025},
url={https://openreview.net/forum?id=W49UjcpGxx}
}

@InProceedings{Liu_2025_CVPR,
    author    = {Liu, Feng and Zhang, Shiwei and Wang, Xiaofeng and Wei, Yujie and Qiu, Haonan and Zhao, Yuzhong and Zhang, Yingya and Ye, Qixiang and Wan, Fang},
    title     = {Timestep Embedding Tells: It's Time to Cache for Video Diffusion Model},
    booktitle = {Proceedings of the IEEE/CVF Conference on Computer Vision and Pattern Recognition (CVPR)},
    month     = {June},
    year      = {2025},
    pages     = {7353-7363}
}

@InProceedings{Huang_2024_CVPR,
    author    = {Huang, Ziqi and He, Yinan and Yu, Jiashuo and Zhang, Fan and Si, Chenyang and Jiang, Yuming and Zhang, Yuanhan and Wu, Tianxing and Jin, Qingyang and Chanpaisit, Nattapol and Wang, Yaohui and Chen, Xinyuan and Wang, Limin and Lin, Dahua and Qiao, Yu and Liu, Ziwei},
    title     = {VBench: Comprehensive Benchmark Suite for Video Generative Models},
    booktitle = {Proceedings of the IEEE/CVF Conference on Computer Vision and Pattern Recognition (CVPR)},
    month     = {June},
    year      = {2024},
    pages     = {21807-21818}
}

@InProceedings{Ma_2024_CVPR,
    author    = {Ma, Xinyin and Fang, Gongfan and Wang, Xinchao},
    title     = {DeepCache: Accelerating Diffusion Models for Free},
    booktitle = {Proceedings of the IEEE/CVF Conference on Computer Vision and Pattern Recognition (CVPR)},
    month     = {June},
    year      = {2024},
    pages     = {15762-15772}
}

@InProceedings{Liu_2025_ICCV,
    author    = {Liu, Jiacheng and Zou, Chang and Lyu, Yuanhuiyi and Chen, Junjie and Zhang, Linfeng},
    title     = {From Reusing to Forecasting: Accelerating Diffusion Models with TaylorSeers},
    booktitle = {Proceedings of the IEEE/CVF International Conference on Computer Vision (ICCV)},
    month     = {October},
    year      = {2025},
    pages     = {15853-15863}
}

@InProceedings{Kahatapitiya_2025_ICCV,
    author    = {Kahatapitiya, Kumara and Liu, Haozhe and He, Sen and Liu, Ding and Jia, Menglin and Zhang, Chenyang and Ryoo, Michael S. and Xie, Tian},
    title     = {Adaptive Caching for Faster Video Generation with Diffusion Transformers},
    booktitle = {Proceedings of the IEEE/CVF International Conference on Computer Vision (ICCV)},
    month     = {October},
    year      = {2025},
    pages     = {15240-15252}
}

@InProceedings{Peebles_2023_ICCV,
    author    = {Peebles, William and Xie, Saining},
    title     = {Scalable Diffusion Models with Transformers},
    booktitle = {Proceedings of the IEEE/CVF International Conference on Computer Vision (ICCV)},
    month     = {October},
    year      = {2023},
    pages     = {4195-4205}
}

@inproceedings{NEURIPS2020_4c5bcfec,
 author = {Ho, Jonathan and Jain, Ajay and Abbeel, Pieter},
 booktitle = {Advances in Neural Information Processing Systems},
 editor = {H. Larochelle and M. Ranzato and R. Hadsell and M.F. Balcan and H. Lin},
 pages = {6840--6851},
 publisher = {Curran Associates, Inc.},
 title = {Denoising Diffusion Probabilistic Models},
 url = {https://proceedings.neurips.cc/paper_files/paper/2020/file/4c5bcfec8584af0d967f1ab10179ca4b-Paper.pdf},
 volume = {33},
 year = {2020}
}

@inproceedings{NEURIPS2022_260a14ac,
 author = {Lu, Cheng and Zhou, Yuhao and Bao, Fan and Chen, Jianfei and LI, Chongxuan and Zhu, Jun},
 booktitle = {Advances in Neural Information Processing Systems},
 doi = {10.52202/068431-0418},
 editor = {S. Koyejo and S. Mohamed and A. Agarwal and D. Belgrave and K. Cho and A. Oh},
 pages = {5775--5787},
 publisher = {Curran Associates, Inc.},
 title = {DPM-Solver: A Fast ODE Solver for Diffusion Probabilistic Model Sampling in Around 10 Steps},
 url = {https://proceedings.neurips.cc/paper_files/paper/2022/file/260a14acce2a89dad36adc8eefe7c59e-Paper-Conference.pdf},
 volume = {35},
 year = {2022}
}

@InProceedings{Cai_2026_CVPR,
    author    = {Cai, Peiliang and Liu, Jiacheng and Xu, Haowen and Wang, Xinyu and Zou, Chang and Zhang, Linfeng},
    title     = {LESA: Learnable Stage-Aware Predictors for Diffusion Model Acceleration},
    booktitle = {Proceedings of the IEEE/CVF Conference on Computer Vision and Pattern Recognition (CVPR)},
    month     = {June},
    year      = {2026},
    pages     = {43300-43309}
}

@misc{song2022denoisingdiffusionimplicitmodels,
      title={Denoising Diffusion Implicit Models}, 
      author={Jiaming Song and Chenlin Meng and Stefano Ermon},
      year={2022},
      eprint={2010.02502},
      archivePrefix={arXiv},
      primaryClass={cs.LG},
      url={https://arxiv.org/abs/2010.02502}, 
}

@inproceedings{
ma2025magcache,
title={MagCache: Fast Video Generation with Magnitude-Aware Cache},
author={Zehong Ma and Longhui Wei and Feng Wang and Shiliang Zhang and Qi Tian},
booktitle={The Thirty-ninth Annual Conference on Neural Information Processing Systems},
year={2025},
url={https://openreview.net/forum?id=KZn7TDOL4J}
}

@inproceedings{
ma2024learningtocache,
title={Learning-to-Cache: Accelerating Diffusion Transformer via Layer Caching},
author={Xinyin Ma and Gongfan Fang and Michael Bi Mi and Xinchao Wang},
booktitle={The Thirty-eighth Annual Conference on Neural Information Processing Systems},
year={2024},
url={https://openreview.net/forum?id=ZupoMzMNrO}
}

@inproceedings{
lei2026budgetconstrained,
title={Budget-Constrained Step-Level Diffusion Caching},
author={Mingkun Lei and Tong Zhao and Liangyu Yuan and Chi Zhang},
booktitle={Forty-third International Conference on Machine Learning},
year={2026},
url={https://openreview.net/forum?id=kN0NMoVWXG}
}

@misc{zhao2026bagbudgetawaregatingdiffusion,
      title={BAG: Budget-Aware Gating for Diffusion Caching}, 
      author={Tong Zhao and Mingkun Lei and Yucheng Han and Chi Zhang},
      year={2026},
      eprint={2608.09231},
      archivePrefix={arXiv},
      primaryClass={cs.CV},
      url={https://arxiv.org/abs/2608.09231}, 
}

@misc{ye2026localmismatchglobalimpact,
      title={From Local Mismatch to Global Impact: Optimizing Cache Reuse Policy for Efficient Diffusion}, 
      author={Xichen Ye and Yifan Wu and Zhikang Xie and Xiangyu Yue and Cheng Jin and Weizhong Zhang},
      year={2026},
      eprint={2608.13043},
      archivePrefix={arXiv},
      primaryClass={cs.AI},
      url={https://arxiv.org/abs/2608.13043}, 
}

@misc{wei2026adaptivehybridcachingefficient,
      title={Adaptive Hybrid Caching for Efficient Text-to-Video Diffusion Model Acceleration}, 
      author={Yuanxin Wei and Lansong Diao and Bujiao Chen and Shenggan Cheng and Zhengping Qian and Wenyuan Yu and Nong Xiao and Wei Lin and Jiangsu Du},
      year={2026},
      eprint={2508.12691},
      archivePrefix={arXiv},
      primaryClass={cs.GR},
      url={https://arxiv.org/abs/2508.12691}, 
}

@misc{ho2022imagenvideohighdefinition,
      title={Imagen Video: High Definition Video Generation with Diffusion Models}, 
      author={Jonathan Ho and William Chan and Chitwan Saharia and Jay Whang and Ruiqi Gao and Alexey Gritsenko and Diederik P. Kingma and Ben Poole and Mohammad Norouzi and David J. Fleet and Tim Salimans},
      year={2022},
      eprint={2210.02303},
      archivePrefix={arXiv},
      primaryClass={cs.CV},
      url={https://arxiv.org/abs/2210.02303}, 
}

@misc{kong2025hunyuanvideosystematicframeworklarge,
      title={HunyuanVideo: A Systematic Framework For Large Video Generative Models}, 
      author={Weijie Kong and Qi Tian and Zijian Zhang and Rox Min and Zuozhuo Dai and Jin Zhou and Jiangfeng Xiong and Xin Li and Bo Wu and Jianwei Zhang and Kathrina Wu and Qin Lin and Junkun Yuan and Yanxin Long and Aladdin Wang and Andong Wang and Changlin Li and Duojun Huang and Fang Yang and Hao Tan and Hongmei Wang and Jacob Song and Jiawang Bai and Jianbing Wu and Jinbao Xue and Joey Wang and Kai Wang and Mengyang Liu and Pengyu Li and Shuai Li and Weiyan Wang and Wenqing Yu and Xinchi Deng and Yang Li and Yi Chen and Yutao Cui and Yuanbo Peng and Zhentao Yu and Zhiyu He and Zhiyong Xu and Zixiang Zhou and Zunnan Xu and Yangyu Tao and Qinglin Lu and Songtao Liu and Dax Zhou and Hongfa Wang and Yong Yang and Di Wang and Yuhong Liu and Jie Jiang and Caesar Zhong},
      year={2025},
      eprint={2412.03603},
      archivePrefix={arXiv},
      primaryClass={cs.CV},
      url={https://arxiv.org/abs/2412.03603}, 
}

@misc{wan2025wanopenadvancedlargescale,
      title={Wan: Open and Advanced Large-Scale Video Generative Models}, 
      author={Team Wan and Ang Wang and Baole Ai and Bin Wen and Chaojie Mao and Chen-Wei Xie and Di Chen and Feiwu Yu and Haiming Zhao and Jianxiao Yang and Jianyuan Zeng and Jiayu Wang and Jingfeng Zhang and Jingren Zhou and Jinkai Wang and Jixuan Chen and Kai Zhu and Kang Zhao and Keyu Yan and Lianghua Huang and Mengyang Feng and Ningyi Zhang and Pandeng Li and Pingyu Wu and Ruihang Chu and Ruili Feng and Shiwei Zhang and Siyang Sun and Tao Fang and Tianxing Wang and Tianyi Gui and Tingyu Weng and Tong Shen and Wei Lin and Wei Wang and Wei Wang and Wenmeng Zhou and Wente Wang and Wenting Shen and Wenyuan Yu and Xianzhong Shi and Xiaoming Huang and Xin Xu and Yan Kou and Yangyu Lv and Yifei Li and Yijing Liu and Yiming Wang and Yingya Zhang and Yitong Huang and Yong Li and You Wu and Yu Liu and Yulin Pan and Yun Zheng and Yuntao Hong and Yupeng Shi and Yutong Feng and Zeyinzi Jiang and Zhen Han and Zhi-Fan Wu and Ziyu Liu},
      year={2025},
      eprint={2503.20314},
      archivePrefix={arXiv},
      primaryClass={cs.CV},
      url={https://arxiv.org/abs/2503.20314}, 
}

@misc{chen2024deltadittrainingfreeaccelerationmethod,
      title={$\Delta$-DiT: A Training-Free Acceleration Method Tailored for Diffusion Transformers}, 
      author={Pengtao Chen and Mingzhu Shen and Peng Ye and Jianjian Cao and Chongjun Tu and Christos-Savvas Bouganis and Yiren Zhao and Tao Chen},
      year={2024},
      eprint={2406.01125},
      archivePrefix={arXiv},
      primaryClass={cs.CV},
      url={https://arxiv.org/abs/2406.01125}, 
}

@InProceedings{Chen_2025_CVPR,
    author    = {Chen, Zhiyuan and Li, Keyi and Jia, Yifan and Ye, Le and Ma, Yufei},
    title     = {Accelerating Diffusion Transformer via Increment-Calibrated Caching with Channel-Aware Singular Value Decomposition},
    booktitle = {Proceedings of the IEEE/CVF Conference on Computer Vision and Pattern Recognition (CVPR)},
    month     = {June},
    year      = {2025},
    pages     = {18011-18020}
}

@InProceedings{Qiu_2025_ICCV,
    author    = {Qiu, Junxiang and Liu, Lin and Wang, Shuo and Lu, Jinda and Chen, Kezhou and Hao, Yanbin},
    title     = {Accelerating Diffusion Transformer via Gradient-Optimized Cache},
    booktitle = {Proceedings of the IEEE/CVF International Conference on Computer Vision (ICCV)},
    month     = {October},
    year      = {2025},
    pages     = {17608-17617}
}

@InProceedings{Nan_2026_CVPR,
    author    = {Nan, Kepan and Zhao, Wangbo and Zhou, Penghao and Li, Jun and Yang, Zhenheng and Yang, Jian and Tai, Ying},
    title     = {Accelerating Autoregressive Video Diffusion via History-Guided Cache and Residual Correction},
    booktitle = {Proceedings of the IEEE/CVF Conference on Computer Vision and Pattern Recognition (CVPR)},
    month     = {June},
    year      = {2026},
    pages     = {43740-43750}
}

@InProceedings{He_2026_CVPR,
    author    = {He, Shaoxuan and Cui, Benlei and Huang, Bukun and Ye, Zhizeng and Sun, Yunyun and Huang, Longtao and Xue, Hui and Yang, Yang and Hong, Haiwen and Tang, Jingqun and Zhao, Zhou},
    title     = {TC-Pade: Trajectory-Consistent Pade Approximation for Diffusion Acceleration},
    booktitle = {Proceedings of the IEEE/CVF Conference on Computer Vision and Pattern Recognition (CVPR)},
    month     = {June},
    year      = {2026},
    pages     = {35768-35778}
}

@misc{sauer2023adversarialdiffusiondistillation,
      title={Adversarial Diffusion Distillation}, 
      author={Axel Sauer and Dominik Lorenz and Andreas Blattmann and Robin Rombach},
      year={2023},
      eprint={2311.17042},
      archivePrefix={arXiv},
      primaryClass={cs.CV},
      url={https://arxiv.org/abs/2311.17042}, 
}

@inproceedings{
song2021scorebased,
title={Score-Based Generative Modeling through Stochastic Differential Equations},
author={Yang Song and Jascha Sohl-Dickstein and Diederik P Kingma and Abhishek Kumar and Stefano Ermon and Ben Poole},
booktitle={International Conference on Learning Representations},
year={2021},
url={https://openreview.net/forum?id=PxTIG12RRHS}
}

@inproceedings{
guo2024animatediff,
title={AnimateDiff: Animate Your Personalized Text-to-Image Diffusion Models without Specific Tuning},
author={Yuwei Guo and Ceyuan Yang and Anyi Rao and Zhengyang Liang and Yaohui Wang and Yu Qiao and Maneesh Agrawala and Dahua Lin and Bo Dai},
booktitle={The Twelfth International Conference on Learning Representations},
year={2024},
url={https://openreview.net/forum?id=Fx2SbBgcte}
}

@article{
ma2025latte,
title={Latte: Latent Diffusion Transformer for Video Generation},
author={Xin Ma and Yaohui Wang and Xinyuan Chen and Gengyun Jia and Ziwei Liu and Yuan-Fang Li and Cunjian Chen and Yu Qiao},
journal={Transactions on Machine Learning Research},
issn={2835-8856},
year={2025},
url={https://openreview.net/forum?id=ntGPYNUF3t},
note={}
}

@inproceedings{
zou2025accelerating,
title={Accelerating Diffusion Transformers with Token-wise Feature Caching},
author={Chang Zou and Xuyang Liu and Ting Liu and Siteng Huang and Linfeng Zhang},
booktitle={The Thirteenth International Conference on Learning Representations},
year={2025},
url={https://openreview.net/forum?id=yYZbZGo4ei}
}

@inproceedings{
puigcerver2024from,
title={From Sparse to Soft Mixtures of Experts},
author={Joan Puigcerver and Carlos Riquelme Ruiz and Basil Mustafa and Neil Houlsby},
booktitle={The Twelfth International Conference on Learning Representations},
year={2024},
url={https://openreview.net/forum?id=jxpsAj7ltE}
}

@inproceedings{ma2026moegps,
title={{MoE-GPS}: Guidlines for Prediction Strategy with Expert Duplication in {MoE} Load Balancing},
author={Haiyue Ma and Zhixu Du and Yiran Chen},
booktitle={NeurIPS 2025 Workshop on Machine Learning for Systems},
year={2025},
url={https://openreview.net/forum?id=VCh5vBSxxb}
}

@InProceedings{Rombach_2022_CVPR,
    author    = {Rombach, Robin and Blattmann, Andreas and Lorenz, Dominik and Esser, Patrick and Ommer, Bj\"orn},
    title     = {High-Resolution Image Synthesis With Latent Diffusion Models},
    booktitle = {Proceedings of the IEEE/CVF Conference on Computer Vision and Pattern Recognition (CVPR)},
    month     = {June},
    year      = {2022},
    pages     = {10684-10695}
}

@InProceedings{Blattmann_2023_CVPR,
    author    = {Blattmann, Andreas and Rombach, Robin and Ling, Huan and Dockhorn, Tim and Kim, Seung Wook and Fidler, Sanja and Kreis, Karsten},
    title     = {Align Your Latents: High-Resolution Video Synthesis With Latent Diffusion Models},
    booktitle = {Proceedings of the IEEE/CVF Conference on Computer Vision and Pattern Recognition (CVPR)},
    month     = {June},
    year      = {2023},
    pages     = {22563-22575}
}

@InProceedings{Wimbauer_2024_CVPR,
    author    = {Wimbauer, Felix and Wu, Bichen and Schoenfeld, Edgar and Dai, Xiaoliang and Hou, Ji and He, Zijian and Sanakoyeu, Artsiom and Zhang, Peizhao and Tsai, Sam and Kohler, Jonas and Rupprecht, Christian and Cremers, Daniel and Vajda, Peter and Wang, Jialiang},
    title     = {Cache Me if You Can: Accelerating Diffusion Models through Block Caching},
    booktitle = {Proceedings of the IEEE/CVF Conference on Computer Vision and Pattern Recognition (CVPR)},
    month     = {June},
    year      = {2024},
    pages     = {6211-6220}
}

@InProceedings{Yin_2024_CVPR,
    author    = {Yin, Tianwei and Gharbi, Micha\"el and Zhang, Richard and Shechtman, Eli and Durand, Fr\'edo and Freeman, William T. and Park, Taesung},
    title     = {One-step Diffusion with Distribution Matching Distillation},
    booktitle = {Proceedings of the IEEE/CVF Conference on Computer Vision and Pattern Recognition (CVPR)},
    month     = {June},
    year      = {2024},
    pages     = {6613-6623}
}

@InProceedings{Du_2025_ICCV,
    author    = {Du, Zhenbang and Fu, Yonggan and Wang, Lifu and Qian, Jiayi and Luo, Xiao and Lin, Yingyan Celine},
    title     = {Fewer Denoising Steps or Cheaper Per-Step Inference: Towards Compute-Optimal Diffusion Model Deployment},
    booktitle = {Proceedings of the IEEE/CVF International Conference on Computer Vision (ICCV)},
    month     = {October},
    year      = {2025},
    pages     = {3001-3010}
}

@article{JMLR:v23:21-0998,
  author  = {William Fedus and Barret Zoph and Noam Shazeer},
  title   = {Switch Transformers: Scaling to Trillion Parameter Models with Simple and Efficient Sparsity},
  journal = {Journal of Machine Learning Research},
  year    = {2022},
  volume  = {23},
  number  = {120},
  pages   = {1--39},
  url     = {http://jmlr.org/papers/v23/21-0998.html}
}

@inproceedings{MLSYS2024_698cfaf7,
 author = {Du, Zhixu and Li, Shiyu and Wu, Yuhao and Jiang, Xiangyu and Sun, Jingwei and Zheng, Qilin and Wu, Yongkai and Li, Ang and Li, Hai Helen and Chen, Yiran},
 booktitle = {Proceedings of Machine Learning and Systems},
 editor = {P. Gibbons and G. Pekhimenko and C. De Sa},
 pages = {224--238},
 title = {SiDA: Sparsity-Inspired Data-Aware Serving for Efficient and Scalable  Large Mixture-of-Experts Models},
 url = {https://proceedings.mlsys.org/paper_files/paper/2024/file/698cfaf72a208aef2e78bcac55b74328-Paper-Conference.pdf},
 volume = {6},
 year = {2024}
}

@misc{lipman2023flowmatchinggenerativemodeling,
      title={Flow Matching for Generative Modeling}, 
      author={Yaron Lipman and Ricky T. Q. Chen and Heli Ben-Hamu and Maximilian Nickel and Matt Le},
      year={2023},
      eprint={2210.02747},
      archivePrefix={arXiv},
      primaryClass={cs.LG},
      url={https://arxiv.org/abs/2210.02747}, 
}

@misc{liu2022flowstraightfastlearning,
      title={Flow Straight and Fast: Learning to Generate and Transfer Data with Rectified Flow}, 
      author={Xingchao Liu and Chengyue Gong and Qiang Liu},
      year={2022},
      eprint={2209.03003},
      archivePrefix={arXiv},
      primaryClass={cs.LG},
      url={https://arxiv.org/abs/2209.03003}, 
}

@misc{singer2022makeavideotexttovideogenerationtextvideo,
      title={Make-A-Video: Text-to-Video Generation without Text-Video Data}, 
      author={Uriel Singer and Adam Polyak and Thomas Hayes and Xi Yin and Jie An and Songyang Zhang and Qiyuan Hu and Harry Yang and Oron Ashual and Oran Gafni and Devi Parikh and Sonal Gupta and Yaniv Taigman},
      year={2022},
      eprint={2209.14792},
      archivePrefix={arXiv},
      primaryClass={cs.CV},
      url={https://arxiv.org/abs/2209.14792}, 
}

@misc{selvaraju2024forafastforwardcachingdiffusion,
      title={FORA: Fast-Forward Caching in Diffusion Transformer Acceleration}, 
      author={Pratheba Selvaraju and Tianyu Ding and Tianyi Chen and Ilya Zharkov and Luming Liang},
      year={2024},
      eprint={2407.01425},
      archivePrefix={arXiv},
      primaryClass={cs.CV},
      url={https://arxiv.org/abs/2407.01425}, 
}

@misc{salimans2022progressivedistillationfastsampling,
      title={Progressive Distillation for Fast Sampling of Diffusion Models}, 
      author={Tim Salimans and Jonathan Ho},
      year={2022},
      eprint={2202.00512},
      archivePrefix={arXiv},
      primaryClass={cs.LG},
      url={https://arxiv.org/abs/2202.00512}, 
}

@misc{luo2023latentconsistencymodelssynthesizing,
      title={Latent Consistency Models: Synthesizing High-Resolution Images with Few-Step Inference}, 
      author={Simian Luo and Yiqin Tan and Longbo Huang and Jian Li and Hang Zhao},
      year={2023},
      eprint={2310.04378},
      archivePrefix={arXiv},
      primaryClass={cs.CV},
      url={https://arxiv.org/abs/2310.04378}, 
}

@misc{shazeer2017outrageouslylargeneuralnetworks,
      title={Outrageously Large Neural Networks: The Sparsely-Gated Mixture-of-Experts Layer}, 
      author={Noam Shazeer and Azalia Mirhoseini and Krzysztof Maziarz and Andy Davis and Quoc Le and Geoffrey Hinton and Jeff Dean},
      year={2017},
      eprint={1701.06538},
      archivePrefix={arXiv},
      primaryClass={cs.LG},
      url={https://arxiv.org/abs/1701.06538}, 
}

@misc{du2026swimbaswitchmambamodel,
      title={Swimba: Switch Mamba Model Scales State Space Models}, 
      author={Zhixu Du and Krishna Teja Chitty-Venkata and Murali Emani and Venkatram Vishwanath and Hai Helen Li and Yiran Chen},
      year={2026},
      eprint={2603.06938},
      archivePrefix={arXiv},
      primaryClass={cs.LG},
      url={https://arxiv.org/abs/2603.06938}, 
}

@inproceedings{NEURIPS2022_a98846e9,
 author = {Karras, Tero and Aittala, Miika and Aila, Timo and Laine, Samuli},
 booktitle = {Advances in Neural Information Processing Systems},
 doi = {10.52202/068431-1926},
 editor = {S. Koyejo and S. Mohamed and A. Agarwal and D. Belgrave and K. Cho and A. Oh},
 pages = {26565--26577},
 publisher = {Curran Associates, Inc.},
 title = {Elucidating the Design Space of Diffusion-Based Generative Models},
 url = {https://proceedings.neurips.cc/paper_files/paper/2022/file/a98846e9d9cc01cfb87eb694d946ce6b-Paper-Conference.pdf},
 volume = {35},
 year = {2022}
}

@inproceedings{NEURIPS2021_49ad23d1,
 author = {Dhariwal, Prafulla and Nichol, Alexander},
 booktitle = {Advances in Neural Information Processing Systems},
 editor = {M. Ranzato and A. Beygelzimer and Y. Dauphin and P.S. Liang and J. Wortman Vaughan},
 pages = {8780--8794},
 publisher = {Curran Associates, Inc.},
 title = {Diffusion Models Beat GANs on Image Synthesis},
 url = {https://proceedings.neurips.cc/paper_files/paper/2021/file/49ad23d1ec9fa4bd8d77d02681df5cfa-Paper.pdf},
 volume = {34},
 year = {2021}
}

@inproceedings{NEURIPS2021_48237d9f,
 author = {Riquelme, Carlos and Puigcerver, Joan and Mustafa, Basil and Neumann, Maxim and Jenatton, Rodolphe and Susano Pinto, Andr\'{e} and Keysers, Daniel and Houlsby, Neil},
 booktitle = {Advances in Neural Information Processing Systems},
 editor = {M. Ranzato and A. Beygelzimer and Y. Dauphin and P.S. Liang and J. Wortman Vaughan},
 pages = {8583--8595},
 publisher = {Curran Associates, Inc.},
 title = {Scaling Vision with Sparse Mixture of Experts},
 url = {https://proceedings.neurips.cc/paper_files/paper/2021/file/48237d9f2dea8c74c2a72126cf63d933-Paper.pdf},
 volume = {34},
 year = {2021}
}

@inproceedings{NEURIPS2022_39235c56,
 author = {Ho, Jonathan and Salimans, Tim and Gritsenko, Alexey and Chan, William and Norouzi, Mohammad and Fleet, David},
 booktitle = {Advances in Neural Information Processing Systems},
 doi = {10.52202/068431-0628},
 editor = {S. Koyejo and S. Mohamed and A. Agarwal and D. Belgrave and K. Cho and A. Oh},
 pages = {8633--8646},
 publisher = {Curran Associates, Inc.},
 title = {Video Diffusion Models},
 url = {https://proceedings.neurips.cc/paper_files/paper/2022/file/39235c56aef13fb05a6adc95eb9d8d66-Paper-Conference.pdf},
 volume = {35},
 year = {2022}
}

@inproceedings{ICLR2025_092c2d45,
 author = {Zhao, Xuanlei and Jin, Xiaolong and Wang, Kai and You, Yang},
 booktitle = {International Conference on Learning Representations},
 editor = {Y. Yue and A. Garg and N. Peng and F. Sha and R. Yu},
 pages = {3296--3319},
 title = {Real-Time Video Generation with Pyramid Attention Broadcast},
 url = {https://proceedings.iclr.cc/paper_files/paper/2025/file/092c2d45005ea2db40fc24c470663416-Paper-Conference.pdf},
 volume = {2025},
 year = {2025}
}

@InProceedings{pmlr-v202-song23a,
  title = 	 {Consistency Models},
  author =       {Song, Yang and Dhariwal, Prafulla and Chen, Mark and Sutskever, Ilya},
  booktitle = 	 {Proceedings of the 40th International Conference on Machine Learning},
  pages = 	 {32211--32252},
  year = 	 {2023},
  editor = 	 {Krause, Andreas and Brunskill, Emma and Cho, Kyunghyun and Engelhardt, Barbara and Sabato, Sivan and Scarlett, Jonathan},
  volume = 	 {202},
  series = 	 {Proceedings of Machine Learning Research},
  month = 	 {23--29 Jul},
  publisher =    {PMLR},
  url = 	 {https://proceedings.mlr.press/v202/song23a.html}
}

@misc{liu2025fasterdiffusiontemporalattention,
      title={Faster Diffusion via Temporal Attention Decomposition}, 
      author={Haozhe Liu and Wentian Zhang and Jinheng Xie and Francesco Faccio and Mengmeng Xu and Tao Xiang and Mike Zheng Shou and Juan-Manuel Perez-Rua and Jürgen Schmidhuber},
      year={2025},
      eprint={2404.02747},
      archivePrefix={arXiv},
      primaryClass={cs.CV},
      url={https://arxiv.org/abs/2404.02747}, 
}
}


\end{document}